\documentclass{article}

    \PassOptionsToPackage{numbers}{natbib}
\usepackage[preprint]{neurips_2025}

\usepackage[utf8]{inputenc} % allow utf-8 input
\usepackage[T1]{fontenc}    % use 8-bit T1 fonts
\usepackage{hyperref}       % hyperlinks
\usepackage{url}            % simple URL typesetting
\usepackage{booktabs}       % professional-quality tables
\usepackage{amsfonts}       % blackboard math symbols
\usepackage{nicefrac}       % compact symbols for 1/2, etc.
\usepackage{microtype}      % microtypography
\usepackage{xcolor}         % colors
\usepackage{graphicx}
\usepackage{epstopdf}
\usepackage{makecell}
\usepackage[version=4]{mhchem}
\usepackage{multirow}
\usepackage{multicol}
\usepackage{array}
\usepackage{subcaption}  
\usepackage{caption}   
\usepackage{amsmath}
\usepackage{amssymb}
\usepackage{mathtools}
\usepackage{amsthm}
\usepackage{subcaption}
\usepackage{wrapfig}
\RequirePackage{algorithm}
\RequirePackage{algorithmic}

\usepackage[capitalize,noabbrev]{cleveref}
\usepackage[most]{tcolorbox}
\usepackage{enumitem}
\usepackage[textsize=tiny]{todonotes}

\NewDocumentCommand{\shuiwang}
{ mO{} }{\textcolor{blue}{\textsuperscript{\textit{Shuiwang Ji}}\textsf{\textbf{\small[#1]}}}}

\usepackage{listings}
\usepackage{xcolor}
\usepackage[scaled]{beramono}
\lstdefinestyle{mystyle}{
  language=Python,
  basicstyle=\rmfamily\small,
  keywordstyle=\color{blue},
  stringstyle=\color{orange},
  commentstyle=\color{gray},
  breaklines=true,
  showstringspaces=false,
  frame=single,
  numbers=none,
  captionpos=b,
  escapeinside={(*@}{@*)}
}
\title{Toward Greater Autonomy in Materials Discovery Agents: Unifying Planning, Physics, and Scientists}

\author{%
  Lianhao Zhou\textsuperscript{$1$}\thanks{Equal contribution}
  \quad 
  Hongyi Ling\textsuperscript{$1$}\footnotemark[1]
  \quad
  Keqiang Yan\textsuperscript{$1$}\footnotemark[1]
  \quad
  \textbf{Kaiji Zhao}\textsuperscript{$2$}\quad
  \textbf{Xiaoning Qian}\textsuperscript{$1,3,4$}\\ 
  \textbf{Raymundo Arróyave}\textsuperscript{$2$}\quad 
  \textbf{Xiaofeng Qian}\textsuperscript{$2,3,5$}\quad
  \textbf{Shuiwang Ji}\textsuperscript{$1$}\thanks{Correspondence to: Shuiwang Ji <sji@tamu.edu>} \\ [5pt]
  \textsuperscript{$1$}Department of Computer Science and Engineering, Texas A\&M University  \\ % [5pt]
  \textsuperscript{$2$}Department of Materials Science and Engineering, Texas A\&M University  \\
 \textsuperscript{$3$}Department of Electrical and Computer Engineering, Texas A\&M University\\
 \textsuperscript{$4$}Computing and Data Sciences, Brookhaven National Laboratory\\
 \textsuperscript{$5$}Department of Physics and Astronomy, Texas A\&M University
}   
  
\begin{document}

\maketitle

\begin{abstract}
We aim at designing language agents with greater autonomy for crystal materials discovery. While most of existing studies restrict the agents to perform specific tasks within predefined workflows, we aim to 
automate workflow planning given high-level goals and scientist intuition. To this end, we propose Materials Agent unifying Planning, Physics, and Scientists, known as MAPPS. MAPPS consists of a Workflow Planner, a Tool Code Generator, and a Scientific Mediator.
The Workflow Planner uses large language models (LLMs) to generate structured and multi-step workflows. The Tool Code Generator synthesizes executable Python code for various tasks, including invoking a force field foundation model that encodes physics. The Scientific Mediator coordinates communications, facilitates scientist feedback, and ensures robustness through error reflection and recovery. By unifying planning, physics, and scientists, MAPPS enables flexible and reliable materials discovery with greater autonomy, achieving a five-fold improvement in stability, uniqueness, and novelty rates compared with prior generative models when evaluated on the MP-20 data.
We provide extensive experiments across diverse tasks to show that MAPPS is a promising framework for autonomous materials discovery.
\end{abstract}

\section{Introduction}

Materials discovery has significant societal impacts across energy, environment, health, and beyond. However, its current pace remains limited by a heavy reliance on trial-and-error wet-lab experiments. Computational methods, including those based on density functional theory (DFT), have substantially accelerated materials discovery over the past several decades. Nevertheless, the computational cost of solving DFT is expensive, making them infeasible for exploring the vast and largely uncharted space of stable materials.
In the past a few years, advances in AI for science has led to a new paradigm for scientific discovery by providing significant speed-ups over traditional DFT based methods. There are predictive models~\citep{choudhary2020joint,yan2022periodic,yan2024space,yan2024complete,choudhary2024jarvis} proposed to predict physical properties of atomistic systems with remarkable efficiency and accuracy, and generative models~\citep{jiao2023crystal,antunes2024crystal,yan2024invariant,zhang2023artificial} that can generate novel and stable materials with desired properties. 
Powered by large language models (LLMs), recent studies have also started exploring how LLM-based AI agents can support autonomous materials discovery~\citep{jia2024llmatdesign,zhang2024comprehensive}.
Currently, most existing studies constrain LLM agents to perform predefined actions specified by human experts, with fixed tasks at each step. In these setups, a fixed discovery pipeline is defined in advance, and LLM agents primarily serve to coordinate AI tools~\citep{zou2025agente}.

%However, these AI methods are designed for specific %tasks with fixed input-output formats. They serve as %tools that assist in discovery but do not act as %human experts capable of initiating, guiding, or %controlling the discovery process. 

Here, we attempt to enable more autonomy in materials discovery agents by making use of the planning capabilities of LLMs. While LLMs' planning capabilities for generic and complex tasks are unclear and still a topic of intensive research and discussions, we attempt to explore their performance in a constrained setting of materials discovery and
particularly in scientific workflow planning. This refers to the construction and adaptation of structured sequences of domain-specific actions designed to solve scientific problems. We show that, while the broader capabilities of LLMs in general-purpose planning remain limited, their emerging ability to perform scientific workflow planning is promising, especially when coupled with human scientist interactions. This focused planning capability opens the door to more autonomous and adaptive agents, enabling systems that can reason about goals, generate workflows, and revise their plans dynamically to achieve scientific objectives. 

% \citep{zhang2023artificial}

Concretely, unlike most existing approaches that constrain agents with predefined workflows tailored to specific tasks, our focus is on enabling agents to plan workflows and reason independently. Instead of prescribing step-by-step procedures, we provide only high-level goals and scientific intuition, allowing agents to determine the sequence of actions required to achieve discovery objectives. In addition, we design our agent system to be physics-informed and include human experts in the loop. This setup enriches the agent’s scientific knowledge beyond textual data, mitigates risk, and allows expert guidance to influence the agent’s decisions.
To this end, we propose Materials Agents unifying Planing, Physics, and Scientists, known as MAPPS, a multi-agent system equipped with a Workflow Planner, a Tool Code Generator, and a Scientific Mediator. These three agents, coupled with human scientists, collaboratively drive materials discovery by planning tasks, generating code, and integrating expert guidance. We show that MAPPS achieves a five-fold improvement in stability, uniqueness, and novelty rates compared with prior generative models when evaluated on the MP-20 data.
We provide extensive experiments across diverse tasks to show that MAPPS is a promising framework for autonomous materials discovery.

\section{Materials Agents Unifying Planning, Physics, and Scientists}

The goal of materials discovery is to discover novel materials structures with desirable physical or chemical properties. Following \citet{yan2024invariant}, we represent each crystal structure as a tuple $\mathbf{M} = (\mathbf{X}, \mathbf{P}, \mathbf{L})$, where $\mathbf{X} = [\mathbf{x}_1, \mathbf{x}_2, \cdots, \mathbf{x}_n] \in \mathbb{R}^{d_x \times n}$ denotes the list of $n$ one-hot representations of atom types in the unit cell,  $\mathbf{P} = [\mathbf{p}_1, \mathbf{p}_2, \cdots, \mathbf{p}_n] \in \mathbb{R}^{3 \times n}$  represents the Cartesian coordinates of the atoms, and $\mathbf{L} = [\boldsymbol{\ell}_1, \boldsymbol{\ell}_2, \boldsymbol{\ell}_3] \in \mathbb{R}^{3 \times 3}$ specifies the lattice matrix containing three basic vectors to describe periodic boundary of the unit cell. 

In this paper, we focus on three types of tasks, including crystal generation, crystal structure prediction, and property-guided generation. Crystal generation is the unconditional generation of stable crystal structures without predefined constraints. Crystal structure prediction aims to generate a stable structure given a specific chemical composition. Property-guided generation seeks to design crystal structures that satisfy desired property criteria, such as a target band gap or formation energy. Rather than generating the final structure in a single step, these tasks can be formulated as sequential decision-making problems, where an agent constructs the crystal structure $\mathbf{M}$ through a series of $T$ actions $(a_1, a_2, \ldots, a_T)$. The process may start from an empty structure or from a candidate retrieved from a database, followed by iterative refinement to achieve the design objective.

\subsection{Different Levels of Autonomy in Science Agent Design} \label{subsec:differentlevel}

Given the complexity of such tasks, it becomes crucial to understand the level of autonomy given to the agents performing them. We define three levels of autonomy for agents in materials science discovery, characterized by the agent’s freedom in planning workflows. 

\textbf{Level 1 – Tool-Executing Agents.}  
The agent performs specific tasks within a fixed, human-designed workflow. It operates as a tool integrator or step executor, typically relying on predefined templates or direct calls to existing tools, such as running a DFT calculation. In this setup, planning freedom is minimal, since agents do not alter the overall workflow or sequence of operations and merely automate individual components.

\textbf{Level 2 – Human-Guided Planning Agents.}
The agent proposes workflows by itself, but with human-provided intuition, constraints, or intermediate goals. For instance, the agent may decompose a complex task into sub-tasks based on the domain knowledge provided by experts. Human feedback or verification helps prune the workflow space, allowing for more flexibility than Level 1 while maintaining scientific plausibility and feasibility.

\textbf{Level 3 – Fully Autonomous Planning Agents.}
The agent has full freedom to design and adapt workflows from scratch, with no predefined sequence or human-imposed constraints. It decides which tools to use, how to combine them, and in what order. While this level enables the highest flexibility and potential for novel discoveries, it poses significant challenges in ensuring workflow validity, reliability, and scientific correctness.

\subsection{Overview of MAPPS}

Most existing agent methods in materials science use LLMs as Level 1 autonomous agents, where the model executes isolated tasks within human-curated workflows or serves as a natural language interface to domain-specific tools. These approaches treat the LLM primarily as a tool user, relying heavily on fixed, expert-designed procedures. While effective in individual components, such Level 1 agents are constrained in their ability to adapt or generalize to new scientific challenges.

To improve the autonomy of LLM agents for science, we aim to move beyond systems where agents act solely as tool executors within fixed, human-designed workflows. Instead, we propose a multi-agent framework MAPPS that achieves Level 2 autonomy by enabling agents to actively design and follow their own workflows. Instead of following predefined steps, the agents construct sequences of actions to solve scientific problems, guided by high-level human input such as goals, constraints, or domain heuristics. This design allows the agents to independently develop both solutions and the necessary tools, leading to better adaptability and scientific creativity.

\begin{figure}[t]
    \centering
    \includegraphics[width=1.0\linewidth]{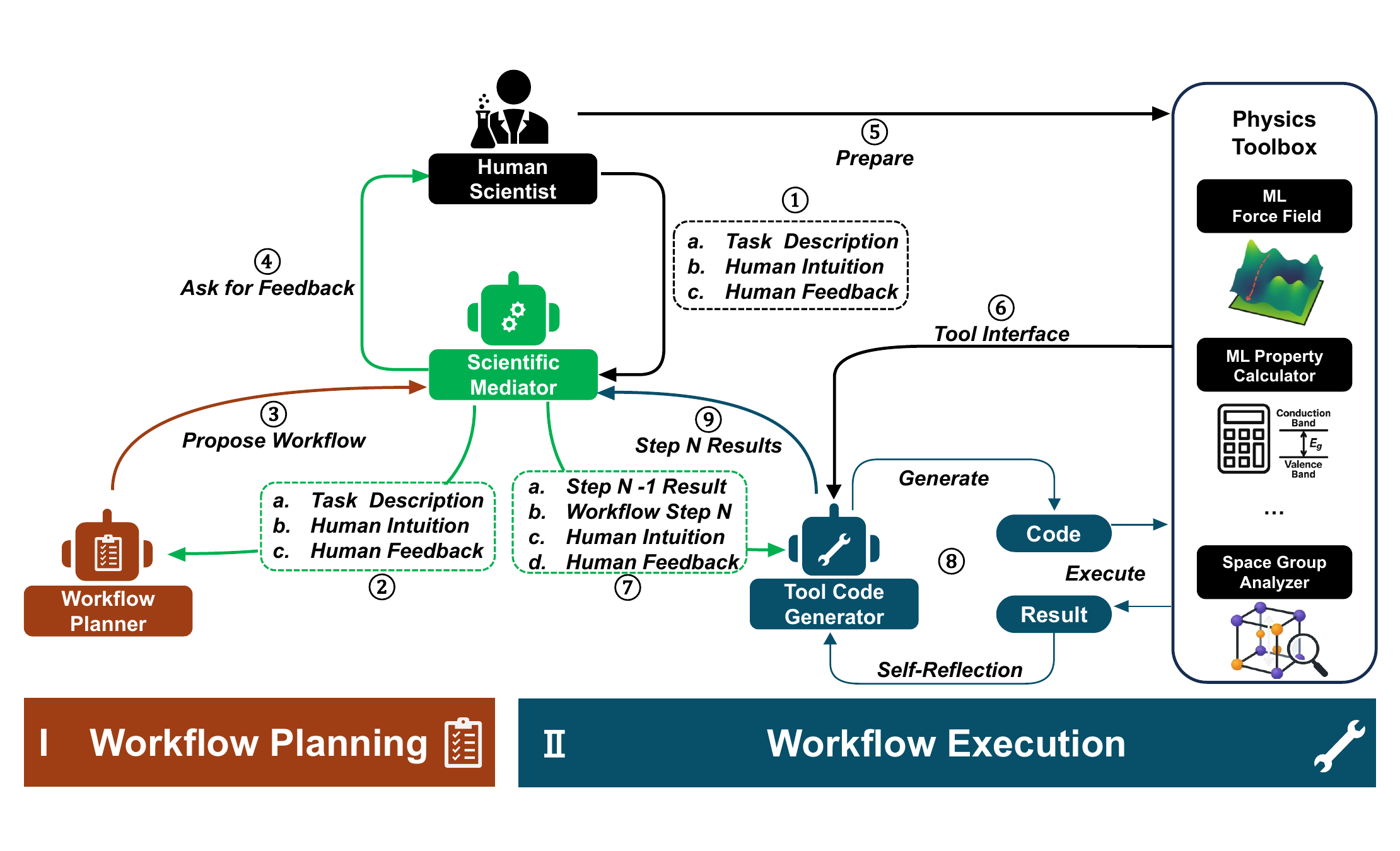}\vspace{-0.6cm}
    \caption{MAPPS Agent Framework. The MAPPS framework consists of three key modules: the Workflow Planner, Tool Code Generator, and Scientific Mediator, which collaboratively drive structure discovery by planning tasks, generating code, and integrating expert guidance. As shown in the figure, the process begins with the human scientist providing a task description, domain intuition, and optional feedback. The Scientific Mediator interprets these inputs and passes them to the Workflow Planner, which proposes a multi-step workflow tailored to the scientific objective. Once the workflow is approved by the human, the Scientific Mediator forwards it to the Tool Code Generator, which translates each workflow step into executable code. This module invokes domain-specific tools from the Physics Toolbox, such as ML Force Fields for structure relaxation, ML Property Calculators for evaluating physical properties, and Space Group Analyzers for symmetry analysis. After execution, results are returned, and the system can optionally engage in self-reflection to detect and correct errors, iteratively improving generated code.}
    \label{fig:MAPPS}
\end{figure}

Specifically, our multi-agent framework integrates three core components, including \textbf{Workflow Planner}, \textbf{Tool Code Generator}, and \textbf{Scientific Mediator}. The Workflow Planner uses an LLM to decompose high-level scientific goals into adaptive, multi-step plans. The Tool Code Generator synthesizes executable code for each step and incorporates physics-based tools, ensuring that outputs are grounded in fundamental physical laws. The Scientific Mediator coordinates communication between agents and humans, maintaining consistency and tracking progress. See Figure~\ref{fig:MAPPS} for an overview of our MAPPS agent framework.

\subsection{Workflow Planner}

To enable autonomous scientific planning, the Workflow Planner is built on a large reasoning model (LRM), which is an advanced LLM with enhanced planning and reasoning capabilities, to generate workflows for different tasks. 
Let $\tau \in \mathcal{T}$ denote a high-level task description, where $\mathcal{T}$ is the space of natural language prompts that specify scientific goals. Given a high-level task description $\tau$, the goal of the Workflow Planner is to generate a workflow consisting of $T$ actionable steps, $\mathbf{A} = (a_1, a_2, \ldots, a_T)$, where each $a_t \in \mathcal{A}$ corresponds to a structured scientific or data processing action, and $\mathcal{A}$ is the space of all executable operations. 
Formally, the workflow generation process can be described as 
\begin{equation}
    \mathbf{A} \sim P_\theta(\mathbf{A} \mid \tau) = \prod_{t=1}^T P_\theta(a_t \mid a_{<t}, \tau),
\end{equation} 
where $P_\theta$ is parameterized by the pretrained LRM and $a_{<t} = (a_1, \ldots, a_{t-1})$ denotes the sequence of previously generated steps.
However, as shown in Section~\ref{subsec:workflowGen}, relying solely on a high-level task description $\tau$ often results in invalid or impractical workflows due to the model's limited domain knowledge. To address this, we introduce an auxiliary input $\iota \in \mathcal{I}$, where $\mathcal{I}$ denotes the space of human-provided intuition, such as domain-specific heuristics. Formally, the refined generation process can be expressed as 
\begin{equation}
    \mathbf{A} \sim P_\theta(\mathbf{A} \mid \tau, \iota) = \prod_{t=1}^T P_\theta(a_t \mid a_{<t}, \tau, \iota).
    \label{eq:workflow}
\end{equation} This refinement guides the model toward more valid plans, while slightly sacrificing the agent’s freedom in exploring arbitrary planning strategies. 

In our implementation, human intuition $\iota$ is provided in a structured prompt with the task description $\tau$. This includes relevant constraints, known physical principles, or useful heuristics, leading to excluding unfeasible operations such as environment setup or model loading from the action space $\mathcal{A}$. Note that the LRM is instructed to output a well-structured workflow containing at most $T=5$ steps.  By injecting expert intuition into the generation process, the Workflow Planner supports Level 2 autonomy, allowing agents to plan more effectively while retaining a human-in-the-loop safeguard. The following template and example demonstrate how human intuition shapes the generated workflow. See Appendix~\ref{appendix:workflow} for the detailed workflow generated by the Workflow Planner. 

\begin{figure}[t]
\begin{tcolorbox}[
    colback=black!10!white, % Background color
    colframe=black,         % Border color
    % boxrule=0.8mm,          % Border thickness
    rounded corners,        % Enables rounded corners
    title= Workflow Planner Prompt Template,
    % fonttitle=\bfseries     % Makes the title bold
]
You are a Workflow Planner. Based on the task requirements and human expert intuition, provide a workflow as a list of necessary steps.
The workflow should contain no more than 5 steps.
Each step must involve data processing — steps such as environment setup, loading models, or loading data are not considered complete steps by themselves. End your output with a note for human approval or feedback.
Each step should be detailed and written on a new line:
\\[1ex]

Step 1:
\\[1ex]
Step 2:

...
\\[1ex]

Task:"task description"
\\[1ex]

Human intuition:"scientist knowledge"

\end{tcolorbox}
\end{figure}
 \subsection{Scientific Mediator}

To support structured interaction between human scientists and autonomous agents, we introduce the Scientific Mediator, a central coordination module that enables a lightweight human-in-the-loop mechanism. While the system is capable of autonomous operation, the Scientific Mediator incorporates human guidance at key decision points to ensure scientific reliability and task relevance.

The process begins when the scientist provides a high-level task description $\tau$. The Scientific Mediator forwards $\tau$ to the Workflow Planner, which generates a structured, multi-step workflow $\mathbf{A} = (a_1, a_2, \ldots, a_T)$. The proposed plan is then returned to the scientist for review and refinement. Once approved, the Scientific Mediator initiates its execution step by step.  At each step $t$, the Scientific Mediator constructs an augmented input context $\xi_t = (a_t, r_{t-1}, \iota_t)$, where $a_t$ is the current action, $r_{t-1}$ denotes the intermediate result from the previous step, and $\iota_t$ is the human intuition at step $t$. This context is sent to the Tool Code Generator to synthesize executable code to perform action $a_t$ and compute the corresponding results $r_t$. 
Before proceeding to step $t+1$, the Scientific Mediator queries the human for approval or feedback on $r_t$, enabling intervention when necessary.
This iterative process maintains a human-in-the-loop mechanism while preserving the autonomy of the system. In this way, the Scientific Mediator plays a crucial role in bridging autonomous agents with human experts, ensuring adaptability and scientific validity.

\subsection{Tool Code Generator}

Given the workflow $\mathbf{A}$ generated by the Workflow Planner, the Tool Code Generator is an autonomous agent responsible for translating each workflow step into executable Python functions. Specifically, at each step $t$, it receives the input context $\xi_t = (a_t, r_{t-1}, \iota_t)$ from the Scientific Mediator and synthesizes a Python function to perform action $a_t$ and compute the result $r_t$.

In addition, the Tool Code Generator operates with a set of domain-specific physics tools denoted as $
\Psi = \{\psi_1, \psi_2, \dots, \psi_n\},$
where each $\psi_i$ represents an individual physics tool. The Tool Code Generator uses a pretrained LRM to generate executable code, integrating these domain-specific physics tools to ensure physically grounded outcomes. For instance, within a crystal structure prediction workflow, the Tool Code Generator integrates a space group analyzer in the initial step to validate symmetry preservation from prototype structures. Subsequently, ML Force Fields (MLFFs) are used to efficiently relax candidate structures towards energetically favorable configurations, significantly reducing computational overhead compared to DFT calculations.

To enhance robustness, the Tool Code Generator includes a self-reflection mechanism. If execution of the generated code results in runtime errors, a diagnostic error signal $e_t$ is generated, triggering the Tool Code Generator itself to revise and regenerate the code. Formally, the initial code generation and self-reflection-based revision processes can be expressed as
\begin{equation}
c_t \sim P_\phi(c_t \mid \xi_t, \Psi) = P_\phi(c_t \mid a_t, r_{t-1}, \iota_t, \Psi),
\label{eq:toolcode}
\end{equation}
where $c_t$ represents the code generated for executing step $a_t$. The revision upon encountering an error is formulated as
\begin{equation}
c_t' \sim P_\phi(c_t' \mid \xi_t, e_t, \Psi),
\end{equation}
where $c_t'$ is the revised code generated for executing step $a_t$.
The result $r_t$ is computed by executing the generated code $c_t$ if no error occurs; otherwise, it is obtained by executing the revised code $c_t'$. See Appendix~\ref{appendix:prompt} for the detailed tool code generation process.

\section{Related Work}

\textbf{Crystal Structure Generation.} Existing methods for 3D crystal generation can be broadly categorized into diffusion-based generative models and language model-based approaches. CDVAE~\citep{cdvae} models the generation of crystal structures by combining variational autoencoders with denoising diffusion. It learns a latent representation of crystals and gradually refines noisy samples into valid structures through a learned reverse diffusion process. DiffCSP~\citep{jiao2023crystal} is a diffusion-based approach specifically designed for crystal structure prediction. It conditions the generation on a given chemical composition and guides the denoising process using surrogate energy models to produce low-energy, stable structures. Mat2Seq~\citep{yan2024invariant} is a language-model-based approach that converts crystal structures into token sequences and trains an autoregressive transformer to generate them. It converts 3D crystal structures into invariant and complete 1D sequences that language models can take as input. CrystaLLM~\citep{antunes2024crystal} trains a language model for crystal generation using text-like representations of crystal structures, i.e., CIF files. However, these generative approaches are designed with specific tasks. They serve as tools that assist in discovery but cannot plan, reason, guide, or control the discovery process.

\textbf{LLM Agents for Science.}
LLM agents are now widely adopted across various scientific domains. In the materials science domain, several systems have been developed to use LLMs for autonomous material discovery. AtomAgents~\citep{ATOMAGENTS} uses a multi-agent framework combining physics-based simulations and multi-modal data integration to design and discover new alloys. OSDA Agent~\citep{OSDA} focuses on zeolite synthesis by integrating molecule generation, quantum evaluations, and reflective feedback to identify suitable organic structure directing agents. LLMatDesign~\citep{jia2024llmatdesign} uses LLMs to translate human instructions into material modifications, applying iterative updates to optimize properties. MatLLMSearch~\citep{gan2025large} demonstrates that pre-trained LLMs, combined with evolutionary search algorithms, can generate stable crystal structures without additional fine-tuning.
Similarly, in other scientific fields, LLM agents have been developed to integrate domain-specific tools within structured workflows~\citep{ghafarollahi2024protagents,drugagent,chatmof,tais,crisprgpt}.
Systems such as ChemCrow~\citep{chemcrow} enable autonomous chemical synthesis by combining LLMs with several chemistry tools. 
Although these systems are promising, LLMs are typically used as tool users who execute predefined steps in workflows designed by human experts and depend heavily on existing domain tools and infrastructure in these systems. As we discussed in Section~\ref{subsec:differentlevel}, this Level 1 usage pattern constrains the autonomy and adaptability of the agent in complex science discovery tasks. 

\textbf{Differences with Prior Work}. MAPPS distinguishes itself from prior agent systems through its ability to autonomously design workflows, implement code, and incorporate intuition and feedback from human experts. We go beyond Level 1 tool-executing agent systems by introducing a Level 2 framework, where agents perform human-guided planning rather than merely executing predefined tasks. Through extensive experiments, we demonstrate that the MAPPS system outperforms Level 1 agent systems, even when those systems follow workflows carefully crafted by human experts.

\section{Experiments}

In this section, we evaluate MAPPS on a diverse range of real-world material discovery tasks, including crystal structure generation, crystal structure prediction, and discovering crystal structures with desired properties. The experimental results demonstrate that our proposed multi-agent framework are able to complete these challenging tasks. Additionally, we present a study in Section~\ref{subsec:workflowGen} to analyze the workflow generations. We conduct our experiments using OpenAI API and a single NVIDIA A100 GPU.

\subsection{Crystal Structure Generation}
\label{subsec:CSG}

\textbf{Setup.} A major goal of materials science is to discover stable and novel crystals. We first evaluate the ability of our proposed multi-agent framework to generate stable crystal structures. We consider two datasets, including MP-20~\citep{jain2013commentary} and Matbench~\citep{dunn2020benchmarking}. MP-20 includes 45,231 stable materials from the Materials Project, covering materials with a maximum of 20 atoms per unit cell and within $0.08$ eV/atom of the convex hull. We follow \citep{gan2025large} to process Matbench. The datasets are used as the retrieval database of our method and the training set of the baselines.  We generate 1,000 candidates on MP-20 and Matbench. 
We evaluate the quality of generated crystal structures using four metrics: validity rate, metastability, stability, and S.U.N rate. Following~\citet{cdvae, court20203, miller2024flowmm}, we compute both structural and compositional validity percentages based on heuristic checks of interatomic distances and charge balance, respectively. For metastability, we adopt the approach of \citet{gan2025large}, using CHGNet~\citep{deng2023chgnet} and M3GNet~\citep{chen2022universal} as surrogate models to estimate the fraction of structures with decomposition energies below thresholds of $0.1$ eV/atom and $0.03$ eV/atom. Stability is further assessed through DFT calculations, where a structure is considered stable if its energy above the convex hull ($E_{\mathrm{hull}}$) is less than $0$ eV/atom. Finally, the S.U.N. rate measures the proportion of structures that are stable, unique, and novel. 

\begin{table}[t]
  \centering
  \fontsize{10}{10}\selectfont  
  \setlength{\tabcolsep}{10pt}
  \caption{Results for crystal structure generation on the MP-20 dataset.}
  \label{tab:MP-20}
  \resizebox{\textwidth}{!}{
  \begin{tabular}{lccccccc}
    \toprule
    Model            & \multicolumn{2}{c}{Validity Rate (\%)}    & \multicolumn{1}{c}{Metastability Rate(\%)}  & Stability Rate (DFT) (\%)             & S.U.N Rate (DFT)(\%) \\
    \cmidrule(lr){2-3} \cmidrule(lr){4-4}
                     & Structural     & Composition     & M3GNet ($E_\mathrm{hull}<0.1$) &            \\
    \midrule
    CDVAE            & 100            & 86.7            & 28.8              & 1.6              & 1.43         \\
    DiffCSP          & 100            & 83.3            & --                & 5.1             & 3.34        \\
    FlowMM           & 96.9           & 83.2            & --                & 4.7             & 2.34        \\
    CrystalTextLLM   & 99.6           & \textbf{95.4}   & \underline{49.8}              & 5.3            & --          \\
    FlowLLM          & 99.9           & 90.8            & --                & \underline{17.8}            & \underline{4.92}        \\
    MAPPS    & \textbf{100}   & \underline{94.0}            & \textbf{95.0}     & \textbf{34.3}   & \textbf{24.9} \\
    \bottomrule
  \end{tabular}
  }
\end{table}

\begin{table}[t]
  \centering
  \fontsize{7}{5}\selectfont  
  \setlength{\tabcolsep}{5pt}
  \caption{Results for crystal structure generation on the Matbench dataset.}
  \label{tab:Matbench}
  \begin{tabular}{lcccccc}
    \toprule
    Model            & \multicolumn{2}{c}{Validity Rate (\%)}    & \multicolumn{3}{c}{Metastability Rate (\%)}  \\
    \cmidrule(lr){2-3} \cmidrule(lr){4-6}
                     & Structural     & Composition     & M3GNet ($E_\mathrm{hull}<0.1$) & CHGNet ($E_\mathrm{hull}<0.1$) & CHGNet($E_\mathrm{hull}<0.03$) &            \\
    \midrule
    MatLLMSearch     & 100            & \textbf{79.4}            & 81.1              & 76.8             & 56.5                 \\
    MAPPS    & \textbf{100}   & 76.9            & \textbf{93.8}     & \textbf{95.9}    & \textbf{84.3}         \\
    \bottomrule
  \end{tabular}
  \vspace{-0.3cm}
\end{table}

\textbf{Baselines.}
We compare MAPPS with the following baseline methods, including (1) CDVAE~\citep{cdvae}, a crystal diffusion variational autoencoder that learn to denoise atomic coordinates and atom types through a diffusion process; (2) DiffCSP~\citep{jiao2023crystal}, which is a diffusion-based generative model that uses a periodic E(3)-equivariant network to jointly generate lattice parameters and fractional atomic coordinates, ensuring symmetry-aware crystal generation; (3) FlowMM~\citep{miller2024flowmm}, a Riemannian flow matching model tailored to crystal symmetries, offering efficient and accurate generation of periodic structures; (4) CrystalTextLLM~\citep{gruver2024fine}, which leverages fine-tuned large language models to generate crystal structures from string-based representations, supporting both unconditional and text-guided generation; (5) FlowLLM~\citep{sriram2024flowllm}, which fine-tunes an LLM to learn an effective base distribution of meta-stable crystals in a text
representation.  and (6) MatLLMSearch~\citep{gan2025large}, which integrates pre-trained LLMs with evolutionary search to iteratively generate and optimize crystal candidates based on structural and property constraints.

\begin{figure}[t]
    \centering
    \includegraphics[width=0.8\linewidth]{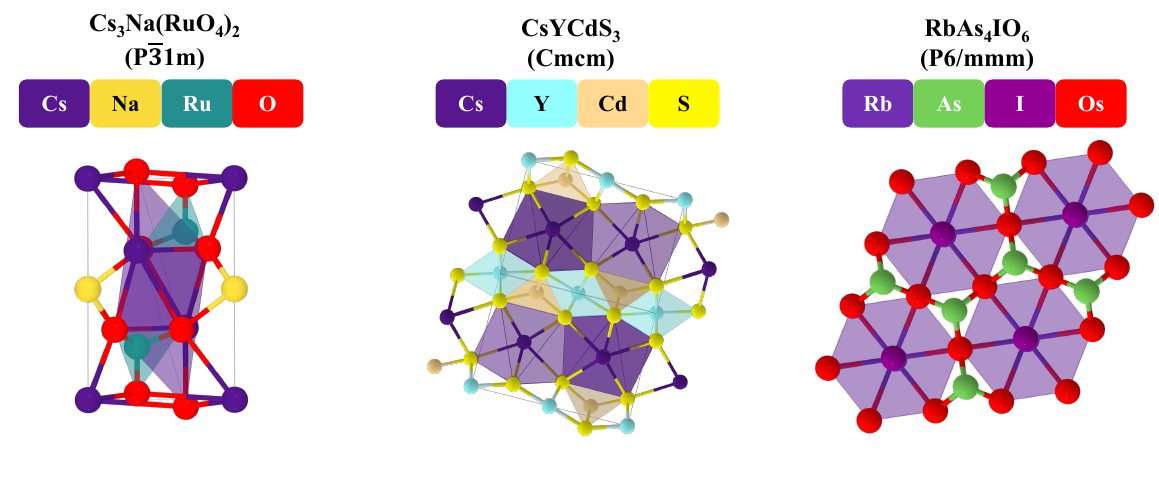}\vspace{-0.3cm}
    \caption{Examples of generated material structures in the Crystal Structure Generation task}
    \label{fig:csg_ex}
\end{figure}

\begin{figure}[t]
    \centering
    \includegraphics[width=0.85\linewidth]{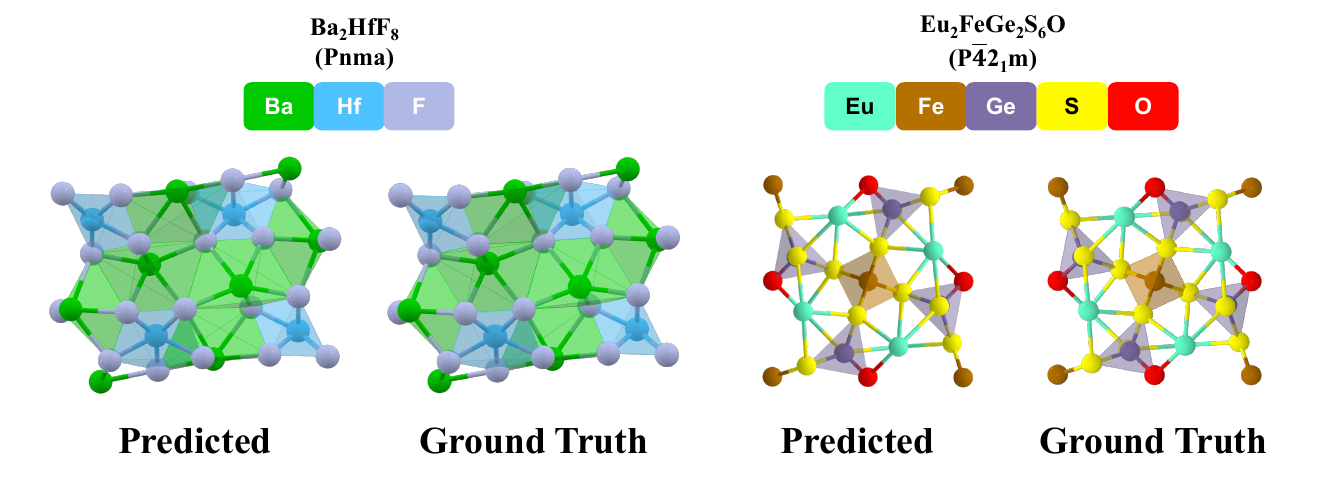}\vspace{-0.2cm}
    \caption{Examples of generated material structures in the Crystal Structure Prediction task}
    \label{fig:csp_ex}
\end{figure}

\textbf{Results.} 
The results in Table~\ref{tab:MP-20} show that MAPPS outperforms several generative model-based baselines, including CDVAE, DiffCSP, FlowMM, and CrystalTextLLM, on the MP-20 dataset. Notably, our approach does not require training any new model. Instead, it successfully extracts the scientific knowledge embedded in pretrained LLMs to solve the crystal structure generation task. This demonstrates that LLMs possess strong capabilities for understanding scientific concepts and facilitating materials discovery.
Moreover, the results in Table~\ref{tab:Matbench} demonstrate that our MAPPS surpasses MatLLMSearch, a recent baseline that combines LLMs with evolutionary algorithms, on the Matbench dataset. This result highlights that LLMs are not merely useful for replacing individual components in an algorithmic pipeline but can serve as central reasoning engines for end-to-end scientific design. We also provide some examples of generated crystal structures in Figure~\ref{fig:csg_ex}.

\subsection{Crystal Structure Prediction}

While the promising results in Section~\ref{subsec:CSG} highlight the effectiveness of our proposed framework, we further evaluate its capability on the crystal structure prediction (CSP) task, which involves predicting the stable structure for a given composition. We use the MP-20 and MPTS-52 dataset\citep{jiao2023crystal}, a challenging benchmark containing $40,476$ structures with up to 52 atoms per unit cell. In addition, we consider the challenge set introduced by~\citet{antunes2024crystal}, which focuses on crystals that have only recently been discovered in the literature, to assess MAPPS's ability to uncover novel structures. Two metrics are used to evaluate the quality of generated crystal structures, namely match rate and RMSE. Match rate measures the ratio of the generated structures that match the ground truth structure determined by the Pymatgen  structure matcher~\citep{pymatgen}. RMSE~\citep{pymatgen}measures the structural differences between the ground truth and matched generated structures.

\textbf{Baselines.}
On the MP-20 and MPTS-52 datasets, we compare MAPPS with several baseline approaches, including language model-based methods such as CrystalLLM~\citep{antunes2024crystal} and Mat2Seq~\citep{yan2024invariant}, as well as diffusion-based methods such as CDVAE~\citep{cdvae} and DiffCSP~\citep{jiao2023crystal}. All baselines are trained using the training sets. To ensure a fair comparison, we also provide our agents with access to the corresponding training data for retrieval. For the challenge set, we compare our method with CrystalLLM~\citep{antunes2024crystal}.  To ensure a fair comparison, we use the same data, which was used by CrystalLLM for training, as our retrieval source. Additionally, all models are evaluated under the same setting of generating a one-shot candidate structure for each input composition.

\textbf{Results.} 
The results in Table~\ref{tab:MP-20_mpts52} show that our approach achieves the highest match rate and the lowest RMSE on both datasets, indicating superior performance of MAPPS on the crystal structure prediction task. On the MP-20 dataset, our method attains a match rate of $63.9\%$ and an RMSE of $0.022$, outperforming all baselines. The performance gains are even more significant on the more challenging MPTS-52 dataset. MAPPS achieves a match rate of $27.6\%$ and an RMSE of $0.097$, significantly outperforming the next-best baseline Mat2Seq by $4.5\%$ in match rate. Notably, on the newly introduced challenge set, MAPPS achieves a match rate of $31.0\%$ and an RMSE of $0.055$, outperforming CrystaLLM by $8.6\%$ in match rate and showing a substantial improvement. These results underscore the effectiveness of MAPPS, demonstrating that Level 2 autonomous agents can generate high-fidelity materials.

\begin{table}[t]
  \centering
  \small
  \caption{Results for crystal structure prediction on three benchmarks, including MP-20, MPTS-52, and the challenge set.}
  \label{tab:MP-20_mpts52}
  \begin{tabular}{lcccccc}
    \toprule
    \textbf{Model} & \multicolumn{2}{c}{\textbf{MP-20}} & \multicolumn{2}{c}{\textbf{MPTS-52}} & \multicolumn{2}{c}{\textbf{Challenge Set}} \\
    \cmidrule(lr){2-3} \cmidrule(lr){4-5} \cmidrule(lr){6-7}
                   & Match Rate & RMSE   & Match Rate & RMSE & Match Rate & RMSE     \\
    \midrule
    
    CDVAE          & 33.9\%     & 0.105  & 5.34\%     & 0.211 & -- & --  \\
    DiffCSP        & 51.5\%     & 0.063  & 12.2\%     & 0.179 & -- & --\\
    CrystaLLM      & 58.7\%     & 0.041  & 19.2\%     & 0.111 & \underline{22.4}\% & \underline{0.090} \\
        Mat2Seq        & \underline{61.3\%}     & \underline{0.040}  & \underline{23.1}\%     & \underline{0.109} & -- & --\\
    MAPPS  & \textbf{63.9\%} & \textbf{0.022} & \textbf{27.6\%} & \textbf{0.097} & \textbf{31.0\%} &\textbf{0.055} \\
    \bottomrule
  \end{tabular}
\end{table}

\begin{table}[!t]
  \centering
  \small
  \caption{Evaluation under Bandgap-Constrained Generation Conditions}
  \label{tab:bandgap}
  \begin{tabular}{lccccc}
    \toprule
    \makecell{Generation Condition}
    & \makecell{Condition Satisfaction (DFT)} 
    & Validity 
    & Uniqueness 
    & Novelty \\
    \midrule
    Bandgap $>$ 3 eV   & 74.6\% & 97.8\% & 96.4\% & 94.8\% \\
    Bandgap $<$ 0.5 eV & 92.2\% & 91.6\% & 98.6\% & 98.0\% \\
    \bottomrule
  \end{tabular}
\end{table}
\vspace{-2ex}

\subsection{Discovering crystal structures with desired properties}
\label{subsec:DCSWDP}

\textbf{Setup.} We further evaluate MAPPS's ability to discover crystal structures with desired electronic properties. Specifically, we focus on generating structures with target bandgap values. We consider two distinct settings, including generating crystals with high bandgaps, defined as bandgap values higher than $3$ eV, and generating crystals with low bandgaps, defined as bandgap values less than $0.5$ eV. For each condition, we generate 500 crystal structures. For retrieval, we use the JARVIS-DFT dataset~\citep{choudhary2020joint}, which contains 61,541 crystal structures along with their corresponding bandgap values. To evaluate the quality of the generated structures, we compute their bandgap values using DFT simulations and report the percentage of generated crystals that meet the high and low bandgap criteria. In addition, we also evaluate the validity, uniqueness, and novelty of the generated structures. See Appendix~\ref{appendix:Bandgap} for the bandgap distributions under two generation settings.

\textbf{Results.} Table~\ref{tab:bandgap} demonstrates that MAPPS can effectively discover crystal structures with target band gap properties. Specifically, under the high band gap setting, $74.6\%$ of the generated crystals have band gap values greater than $3$ eV, while under the low band gap setting, $92.2\%$ of the generated crystals have band gap values below $0.5$ eV. Moreover, the generated structures show high quality, achieving over $90\%$ validity, uniqueness, and novelty.

\begin{table}[!t]
  \centering
  \small
  \caption{Validity (\%) and workflow length under CSP, CSD, and CSG tasks}
  \label{tab:llm-validity}
  \begin{tabular}{cccccccc}
    \toprule
    \textbf{LLM} 
    & \multicolumn{3}{c}{\makecell{Validity \\ W/ Human Intuition}} 
    & \multicolumn{3}{c}{\makecell{Validity \\ W/O Human Intuition}} 
    & \makecell{Avg \\ Workflow Length} \\
    \cmidrule(lr){2-4} \cmidrule(lr){5-7}
    & {CSP} & {CSD} & {CSG} & {CSP} & {CSD} & {CSG} & {} \\
    \midrule
    GPT-4o-mini & 0\%   & 0\%   & 10\%  & 0\% & 0\% & 0\% & 5.0 \\
    GPT-4o      & 10\%  & 30\%  & 40\%  & 0\% & 0\% & 0\% & 5.0 \\
    O3-mini     & 60\%  & 60\%  & 100\% & 0\% & 0\% & 0\% & 4.37 \\
    \bottomrule
  \end{tabular}
\end{table}
\vspace{-2ex}

\subsection{Can current LLMs achieve level 3?}
\label{subsec:workflowGen}
In this subsection, we investigate whether current LLMs are capable of achieving Level 3 autonomy. As we defined in Section~\ref{subsec:differentlevel}, Level 3 autonomous agents have complete freedom to design workflows from scratch without any human-imposed constraints. To evaluate this capability, we assess the performance of three LLMs, namely GPT-4o-mini, GPT-4o, and O3-mini, across three materials discovery tasks, including crystal structure prediction (CSP), crystal structure design (CSD), and crystal structure generation (CSG). Each model is tested under two settings, one where human intuition is provided as guidance and one where it is not. We use the validity rate as our primary evaluation metric, which measures whether the generated workflows are feasible and executable. As shown in Table~\ref{tab:llm-validity}, the recent large reasoning model O3-mini achieves significantly higher validity when guided by human intuition, reaching $60\%$ on both CSP and CSD and $100\%$ on CSG. In comparison, GPT-4o reaches only $10\%$ to $40\%$ validity across these tasks, and GPT-4o-mini performs poorly with validity near zero. Notably, without human guidance, all models fail to produce valid workflows in all three tasks. This finding underscores the importance of expert guidance, even for advanced LLMs, and highlights the challenges of achieving true Level 3 autonomy. Additionally, we observe that despite being prompted to generate no more than five steps, many LLMs consistently produce five-step workflows, often including useless or redundant steps. In contrast, only advanced reasoning models like O3-mini are able to omit unnecessary steps and generate concise, valid workflows with an average length of $4.37$ steps.

\section{Summary and Outlook}

In this work, we introduce MAPPS, a multi-agent framework that unifies Planning, Physics, and Scientists to enable more autonomous materials discovery. 
MAPPS consists of three key components: a Workflow Planner that decomposes high-level scientific goals into actionable steps, a Tool Code Generator that synthesizes executable code grounded in physics, and a Scientific Mediator that incorporates human intuition and feedback throughout the discovery process. 
Extensive experimental results show that our proposed MAPPS system is better than Level 1 agent systems. Current limitations of MAPPS include: (1) MAPPS currently has human expert in the loop, and does not reach full autonomous agent system which is Level 3; and (2) we currently focus on materials discovery tasks, while extensions to other domains such as molecules, polymers, or other systems remain underexplored. Looking forward, level 3 autonomy where agents independently design and adapt workflows without human input, offers exciting potential for scientific discovery. Realizing this vision will require advances in fundamental large language model ability, scientific reasoning, tool code generation, and self-verification. We view MAPPS as a stepping stone toward this goal and leave full Level 3 autonomy as an exciting direction for future work. The discovery of novel materials may have both positive and negative societal impacts relevant to this work.

\section*{Acknowledgments}

SJ acknowledges support from the National Institutes of Health under grant U01AG070112, ARPA-H under grant 1AY1AX000053, and National Science Foundation under grant MOMS-2331036. XFQ acknowledges support from the National Science Foundation under grants CMMI-2226908 and DMR-2103842, as well as partial support by the donors of ACS Petroleum Research Fund under grant \#65502-ND10. RA acknowledges support from the Army Research Laboratory under Cooperative Agreement Number W911NF-22-2-0106 (BIRDSHOT Center) and National Science Foundation under Grant 2119103. XNQ acknowledges partial support from the National Science Foundation under grants SHF-2215573 and IIS-2212419, as well as the Biological and Environmental Research~(BER) program in the US Department of Energy (DOE) Office of Science under project B\&R\# KP1601017 and FWP\#CC140.

%\newpage

\bibliographystyle{unsrtnat}
\bibliography{llmagent,dive,kaiji}

%%%%%%%%%%%%%%%%%%%%%%%%%%%%%%%%%%%%%%%%%%%%%%%%%%%%%%%%%%%%
\newpage
\appendix

\section{Experimental Details}

\subsection{Workflow Planning for Crystal Structure Prediction}\label{appendix:workflow}

The workflow example in the box below is generated by the \textbf{Workflow Planner}, which maps the task description $\tau$ and human intuition $\iota$ to a multi-step action sequence $\mathbf{A} = (a_1, a_2, \ldots, a_T)$. This process follows Equation~\ref{eq:workflow} in the main text:
\vspace{-2ex}
\begin{equation}
\mathbf{A} \sim P_\theta(\mathbf{A} \mid \tau, \iota) = \prod_{t=1}^{T} P_\theta(a_t \mid a_{<t}, \tau, \iota).
\end{equation}

In this example, we provide the following inputs:

\textbf{Task Description} ($\tau$): ``Please predict the stable structure for $\mathrm{Ba_2Fe_2F_9}$.''

\textbf{Human Intuition} ($\iota$): ``1. Recent studies commonly employ Machine Learning force fields as an alternative to Density Functional Theory for optimizing structures and calculating energies. 2. Similar chemical compositions might have similar stable structural prototypes. A dataset is available at \{path\}, providing various structure prototypes.''

The output is the following workflow $\mathbf{A}$:
\vspace{-2ex}
\begin{tcolorbox}[
    colback=black!10!white,
    colframe=black,
    rounded corners,
    title=Workflow Example: Crystal Structure Prediction for $\mathrm{Ba_2Fe_2F_9}$,
]
Step 1: Query the structural database for crystal structures with chemical compositions or reduced formulas similar to $\mathrm{Ba_2Fe_2F_9}$ to identify promising structural prototypes.

Step 2: Use the retrieved similar prototypes to generate initial candidate structures specifically for $\mathrm{Ba_2Fe_2F_9}$, ensuring a diverse set of likely configurations based on known stable arrangements.

Step 3: Optimize these candidate structures using Machine Learning force fields, ensuring the minimization of energy and refinement of lattice and atomic positions.

Step 4: Calculate the total energies for the optimized candidates and compare their stability; the structure with the lowest energy is identified as the most probable stable configuration.

Step 5: Validate the selected structure by cross-referencing with available experimental or high-accuracy computational data, if available, to confirm its stability and consistency with known behavior for similar compounds.

Please review and provide feedback or suggest revisions to the workflow.
\end{tcolorbox}

\subsection{Tool Code Generation}\label{appendix:prompt}

For each workflow step $a_t \in \mathbf{A}$, the \textbf{Scientific Mediator} constructs the input context $\xi_t = (a_t, r_{t-1}, \iota_t)$,
where $a_t$ denotes the current step description, $r_{t-1}$ is the result of the previous step, and $\iota_t$ is the domain-specific expert intuition for step $t$. This input context is passed to the Tool Code Generator, along with a collection of physics tools $\Psi = \{\psi_1, \psi_2, \ldots\}$ (e.g., \textbf{CHGNetCalculator}, \textbf{pymatgen}, \textbf{ASE}), which represent the available modeling and simulation environments. The Tool Code Generator synthesizes executable code $c_t$ based on this context, following Equation~\ref{eq:toolcode} in the main text:

\begin{equation}
c_t \sim P_\phi(c_t \mid \xi_t, \Psi) = P_\phi(c_t \mid a_t, r_{t-1}, \iota_t, \Psi).
\end{equation}

Below, we provide a concrete example for Step 3. The prompt template used for the Tool Code Generator is shown below.
\begin{tcolorbox}[
    colback=black!10!white,
    colframe=black,
    rounded corners,
    title=Tool Code Generator Prompt Template,
]
You are a Tool Code Generator. Based on the following information (last step result, current workflow step, and expert intuition),
please propose complete and executable Python code. The code must define exactly one unique function named 'stepX' (e.g., step1, step2, etc.).
All file paths used in the code must be absolute paths. Save the code result files in a folder and output the folder's path.
Do not print any other information. Return the processing result.
The function must include a descriptive Python comment at the top explaining its purpose and functionality.
\\[2ex]
Output a JSON object in the following format:
\\[1ex]
{
  "function name": "stepX",\\[1ex]
  "code": "Complete Python code"
}
\end{tcolorbox}

The structured inputs to the Tool Code Generator are:

\textbf{Action Description} ($a_3$): ``Optimize candidate structures using Machine Learning force fields, ensuring the minimization of energy and refinement of lattice and atomic positions.''

\textbf{Previous Step Result} ($r_2$): A folder path \texttt{'candidates'} containing candidate structures in 'cif' format.

\textbf{Human Intuition} ($\iota_3$): ``\texttt{CHGNetCalculator} is a machine learning force field that approximates DFT-level accuracy and can be used to efficiently optimize lattice and atomic positions.''

\textbf{Physics Tool Set} ($\Psi$): Includes \texttt{CHGNetCalculator}, \texttt{ASE}, \texttt{pymatgen}.

The following Python function is synthesized by the Tool Code Generator to implement $a_3$.

\begin{lstlisting}[caption={Generated code for Step3}]
def step3(chemical_composition):
    """
    Optimize candidate crystal structures using CHGNetCalculator.
    This function reads candidate cif structures from the folder
    'candidates', assigns CHGNetCalculator as the ASE calculator,
    and optimizes both lattice and atom positions using ExpCellFilter
    and the BFGS optimizer. Optimized structures are saved to
    'optimized_candidates', and the path to this folder is returned.
    """
    import os
    from ase.io import read, write
    from ase.optimize import BFGS
    from ase.constraints import ExpCellFilter
    from chgnet.model import CHGNetCalculator

    input_dir = 'candidates'
    output_dir = 'optimized_candidates'
    os.makedirs(output_dir, exist_ok=True)
    calc = CHGNetCalculator()
    for filename in os.listdir(input_dir):
        if filename.lower().endswith('cif'):
            file_path = os.path.join(input_dir, filename)
            try:
                atoms = read(file_path)
                atoms.set_calculator(calc)
                ecf = ExpCellFilter(atoms)
                opt = BFGS(ecf, logfile=None)
                opt.run(steps=100)
                write(os.path.join(output_dir, filename), atoms)
            except Exception as e:
                print(f'Error processing {file_path}: {e}')
    return output_dir
\end{lstlisting}

This code represents a successful instantiation of the Tool Code Generator output at Step 3, enabling structure relaxation with physics-informed ML force fields.

\section{Bandgap Distribution under Targeted Generation}\label{appendix:Bandgap}
\begin{figure}[t]
  \centering

  % -- Left figure with manual (a)
  \begin{subfigure}[t]{0.45\textwidth}
    \centering
    \includegraphics[width=\linewidth]{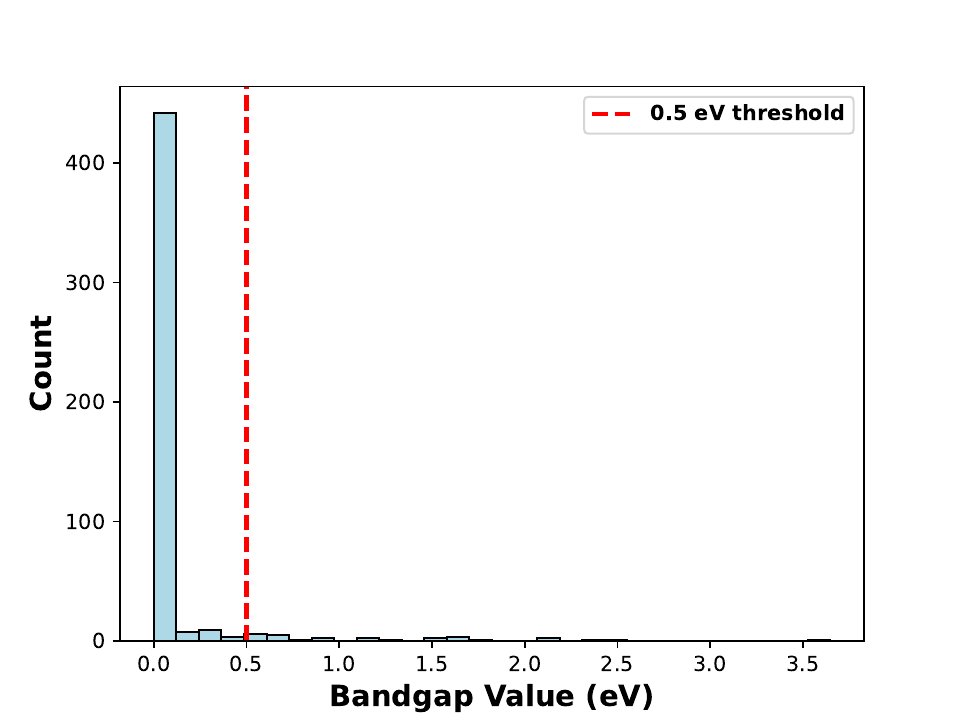}
    \caption*{(a) low-bandgap generation}
  \end{subfigure}
  \hfill
  % -- Right figure with manual (b)
  \begin{subfigure}[t]{0.45\textwidth}
    \centering
    \includegraphics[width=\linewidth]{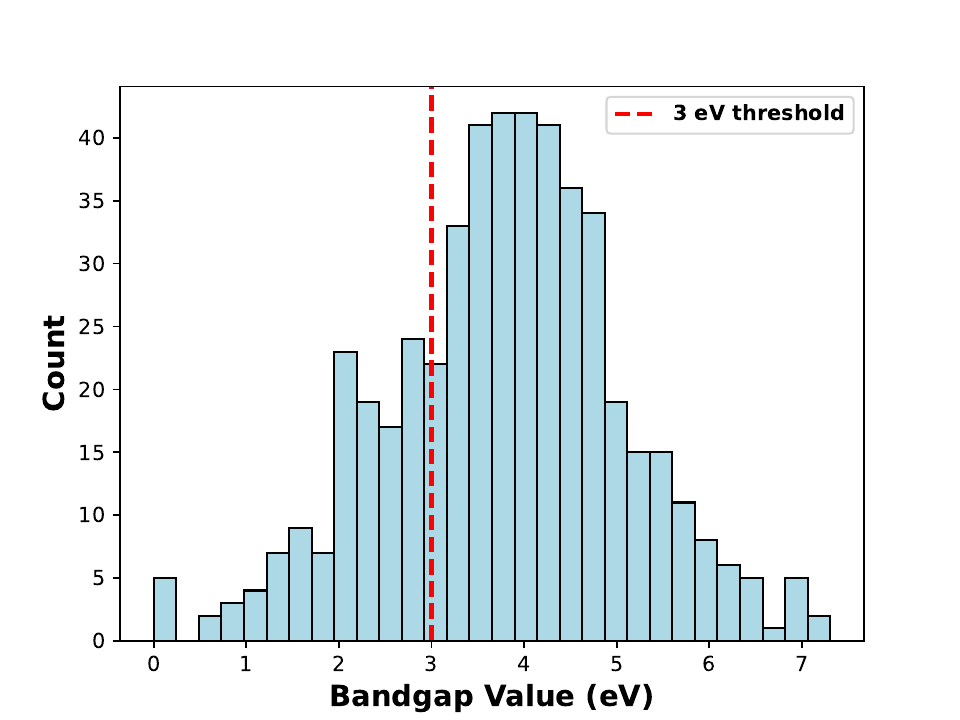}
    \caption*{(b) high-bandgap generation}
  \end{subfigure}

  \caption{DFT-computed bandgap distributions under two generation settings.}
  \label{fig:bandgap}
\end{figure}

In Figure~\ref{fig:bandgap}, we show histograms of DFT-computed bandgaps for structures generated under two targeted conditions in Section~\ref{subsec:DCSWDP}: low-bandgap generation (bandgap $<$ 0.5 eV) and high-bandgap generation (bandgap $>$ 3 eV). Each histogram is computed over 500 generated crystal structures. These plots illustrate how well the generated structures satisfy the intended electronic constraints and how the bandgap distributions differ under each target setting.
Under the low-bandgap setting, a significant proportion of the generated structures exhibit bandgap values below the 0.5 eV threshold, demonstrating the model's ability to synthesize narrow-gap materials such as semimetals or small-gap semiconductors. Conversely, in the high-bandgap setting, the bandgap distribution is shifted toward larger values, with many structures achieving bandgaps greater than 3 eV, indicating the successful generation of wide-gap insulating candidates.

\section{Evaluation Metric Details}\label{appendix:evaluation}

To evaluate the quality of generated crystal structures, we adopt a set of metrics covering structural validity, compositional correctness, thermodynamic stability, uniqueness, novelty, and accuracy of property prediction. Unless otherwise specified, all metrics are computed based on structures post-processed and relaxed by DFT or ML-based surrogates.

\noindent\textbf{Structural Validity.} 
A structure is considered structurally valid if all pairwise interatomic distances are greater than or equal to 0.5~\AA{} and the unit cell volume is no less than 0.1~\AA$^3$. This ensures that generated structures are physically meaningful and free of atom overlaps or degenerate geometries.

\noindent\textbf{Compositional Validity.} 
We assess the physical plausibility of compositions using SMACT~\cite{davies2019smact}, which verifies charge neutrality and electronegativity balance. A crystal is considered compositionally valid if it passes both checks.

\noindent\textbf{Stability.} 
We define a crystal as stable if its DFT-calculated energy above the convex hull is below 0.0~eV/atom and it contains at least two unique elements.

\noindent\textbf{Uniqueness.} 
To measure diversity, we compute the fraction of stable crystals that are mutually unique. Uniqueness is determined via all-to-all structural comparison using the \texttt{StructureMatcher} class from \texttt{pymatgen}. Two crystals are considered duplicates if they match under symmetry-preserving tolerances on lattice, angles, and atomic coordinates.

\noindent\textbf{Novelty.} 
A crystal is considered novel if it does not match any existing structure in the original dataset, again based on the \texttt{StructureMatcher}. This ensures the generated structures are not trivial rediscoveries.

\noindent\textbf{Match Rate.} 
For crystal structure prediction (CSP) tasks, we compute the match rate, defined as the percentage of generated structures that match the ground-truth structure for the given composition, determined using \texttt{StructureMatcher}.

\noindent\textbf{RMSE.} 
We also report the root mean square error (RMSE) between the fractional coordinates of matching atoms in predicted and true structures, after alignment via symmetry operations and cell transformation. RMSE provides a fine-grained measure of geometric fidelity.

\section{DFT calculations}\label{appendix:dft}
First-principles density functional theory (DFT)~\cite{hohenberg1964density, kohn1965self} calculations were performed using the Vienna Ab initio Simulation Package (VASP)~\cite{kresse1996efficient}. For stability and S.U.N rate evaluation in Section~\ref{subsec:CSG}, the Perdew-Burke-Ernzerhof (PBE)~\cite{perdew1996generalized} form of the exchange-correlation functional within the generalized gradient approximation (GGA)~\cite{langreth1983beyond} was employed. To ensure consistency with the MP-20 dataset, all input settings were generated using the MPRelaxSet class. We determine the DFT energy above hull for the relaxed structures against the Matbench Discovery convex hull\cite{riebesell2023matbench}. For the evaluation in Section~\ref{subsec:DCSWDP}, to maintain consistency with the JARVIS-DFT dataset\cite{choudhary2020joint}, the vdW-DF-OptB88 functional~\cite{klimevs2009chemical} was used and the input settings were modified, with the atomic force convergence criterion set to 0.001~eV~$\text{\AA}^{-1}$ and the convergence criterion for electronic self-consistent calculations set to $10^{-7}$~eV.

\end{document}